# Possibilistic Answer Set Programming Revisited

**Kim Bauters**[*], **Steven Schockaert**[†]
Dept. of Applied Mathematics
and Computer Science
Universiteit Gent
Krijgslaan 281
9000 Gent, Belgium

**Martine De Cock**[‡]
Institute of Technology
University of Washington
1900 Commerce Street
WA-98402 Tacoma, USA

**Dirk Vermeir**
Dept. of Computer Science
Vrije Universiteit Brussel
Pleinlaan 2
1050 Brussel, Belgium

## Abstract

Possibilistic answer set programming (PASP) extends answer set programming (ASP) by attaching to each rule a degree of certainty. While such an extension is important from an application point of view, existing semantics are not well-motivated, and do not always yield intuitive results. To develop a more suitable semantics, we first introduce a characterization of answer sets of classical ASP programs in terms of possibilistic logic where an ASP program specifies a set of constraints on possibility distributions. This characterization is then naturally generalized to define answer sets of PASP programs. We furthermore provide a syntactic counterpart, leading to a possibilistic generalization of the well-known Gelfond-Lifschitz reduct, and we show how our framework can readily be implemented using standard ASP solvers.

## 1 Introduction

Answer set programming (ASP) is a form of declarative programming based on the stable model semantics (Gelfond and Lifzchitz, 1988). ASP has proven successful as an elegant formalism for commonsense reasoning in discrete domains and to encode combinatorial problems in a purely declarative way. Possibilistic logic (PL), introduced in (Dubois et al., 1994), emanated from possibility theory (Zadeh, 1978) and offers a sound and complete logic system for representing (partial) ignorance or uncertainty in a non-probabilistic way.

Possibilistic answer set programming (PASP) (Nicolas et al., 2006) unites ASP and PL and provides a single framework that supports declarative programming under uncertainty. In PASP, a certainty value is attached to each rule of an ASP program. The certainty of a particular conclusion of that program is then given by the lowest certainty of the rules that were used, i.e. the strength of a conclusion is determined by the weakest piece of information involved. However, for programs containing negation-as-failure, the semantics proposed in (Nicolas et al., 2006) often lead to counterintuitive results. For example, consider the PASP program $P_1$:

$$\mathbf{1:} \; concertBooked \leftarrow \qquad (1)$$
$$\mathbf{1:} \; longDrive \leftarrow concertBooked, not\ canceled \qquad (2)$$
$$\mathbf{0.2:} \; canceled \leftarrow . \qquad (3)$$

This program encodes that we are certain that we booked a concert and we are certain that we have a long drive ahead of us, unless the concert is canceled. The last rule encodes that the concert is indeed cancelled, although this information comes from an unreliable source, hence a low certainty degree is attached to this rule.

The approach from (Nicolas et al., 2006) consists of first determining the classical answer sets, ignoring the certainty values, and subsequently adding certainties to atoms in these answer sets based on which rules are needed to support them. The program above only has one classical answer set, namely $\{concertBooked, canceled\}$. After adding the certainty degrees, we find the possibilistic answer set

$$\{concertBooked^1, canceled^{0.2}\}$$

indicating that we are entirely certain that we booked the concert and that we are somewhat certain that the concert is canceled. As such, we lose the valuable information that the certainty of the third rule is limited, and thus completely ignore the second rule. Nonetheless, it seems desirable to derive that $longDrive$ may hold, with some limited certainty, unless it is completely certain that we can prove $canceled$. The intuitive meaning of '$not\ a$' is then given by "unless it is certain that 'a' holds". However, this behavior cannot be obtained by adding weights to classical answer sets.

[*]Funded by a joint FWO project.
[†]Postdoctoral fellow of the FWO.
[‡]On leave from Universiteit Gent.

In this paper we take a principled approach to uncover an appropriate semantics for PASP. To this end, we first reveal an important link between ASP and PL, leading to a novel characterization of the stable model semantics. In this characterization, the rules of an ASP program impose constraints on possibility distributions. The possibility distributions that satisfy all these constraints then correspond with the answer sets of the original ASP program. This characterization is then extended in a natural way to cover PASP programs. The resulting semantics will be such that the unique possibilistic answer set of $P_{intro}$ is given by

$$\{concertBooked^1, longDrive^{0.8}, canceled^{0.2}\}.$$

Because our certainty that the concert is canceled is so low, i.e. because we have failed to prove with a high certainty that the concert is canceled, we still derive with a reasonably high certainty that we have a long drive ahead of us.

The remainder of this paper is structured as follows. In Section 2 we start by introducing some background on ASP, PL and PASP. Subsequently, in Section 3 we investigate the links between ASP and PL, leading to an alternative definition of answer sets, in terms of possibilistic logic. Next, we extend this definition in Section 4 to the general case of possibilistic ASP. Because these definitions are formulated at the semantic level, they cannot readily be implemented. To cope with this, we present a syntactic counterpart in Section 5, introducing a possibilistic generalization of the Gelfond-Lifschitz reduct. In Section 6 we then demonstrate how possibilistic answer sets can be computed using existing ASP solvers, thus verifying the applicability of our new approach. Finally, we provide an overview of related work and present some conclusions.

## 2 Preliminaries

### 2.1 Answer Set Programs (ASP)

Let $\mathcal{A}$ be a finite set of atoms. We also consider the special atom $\top$ (or true). A *naf-atom* is an atom preceded by *not* which we call negation-as-failure. Intuitively, *not a* is true when we cannot prove $a$.

A *normal rule* is an expression of the form $a_0 \leftarrow a_1, ..., a_m, not\ a_{m+1}, ..., not\ a_n$ with $a_i$, $0 \leq i \leq n$, an atom of $\mathcal{A}$. We call $a_0$ the *head* of the rule and $a_1, ..., a_m, not\ a_{m+1}, ..., not\ a_n$ the *body* of the rule. If no occurrences of *not* appear in a rule (i.e. $m = n$), the rule is called definite. A *normal (resp. definite) program* $P$ is a finite set of normal (resp. definite) rules. A rule of the form $a_0 \leftarrow$ is called a *fact* and is used as a shorthand for $a_0 \leftarrow \top$. The *Herbrand base* $\mathcal{B}_P$ of a normal program $P$ is the set of atoms appearing in $P$. An *interpretation* $I$ of $P$ is any set of atoms $I \subseteq \mathcal{B}_P$. The set of interpretations $\Omega$ of a normal program $P$ is given by $2^{\mathcal{B}_P}$.

The satisfaction relation $\models$ is defined for an interpretation $I$ as $I \models a$ iff $a = \top$ or $a \in I$. For an interpretation $I$ and $A$ a set of atoms, we define $I \models A$ iff $\forall a \in A \cdot I \models a$. An interpretation $I$ is a *model* of a definite rule $r = a_0 \leftarrow a_1, ..., a_m$, denoted $I \models r$, if $I \models a_0$ or $I \not\models \{a_1, ..., a_m\}$. An interpretation $I$ of a definite program $P$ is a *model* of $P$ iff $\forall r \in P \cdot I \models r$.

Answer sets are defined using the *immediate consequence operator* $T_P$ of a definite program $P$, defined for a set of atoms $I$ as:

$$T_P(I) = I \cup \{a_0 \mid (a_0 \leftarrow a_1, ..., a_m) \in P \\ \land (I \models \{a_1, ..., a_m\})\}$$

We use $P^\star$ to denote the fixpoint which is obtained by repeatedly applying $T_P$ starting from the empty interpretation, i.e. the least fixpoint of $T_P$ w.r.t. set inclusion. An interpretation $I$ is called an *answer set of a definite program* $P$ iff $I = P^\star$. Answer sets of normal programs are defined using the Gelfond-Lifschitz *reduct* $P^I$ of a normal program $P$ and interpretation $I$, defined as

$$P^I = \{a_0 \leftarrow a_1, ..., a_m \mid (\{a_{m+1}, ..., a_n\} \cap I = \emptyset) \\ \land (a_0 \leftarrow a_1, ..., a_m, not\ a_{m+1}, ..., not\ a_n) \in P\}.$$

An interpretation $I$ is then called an *answer set of the normal program* $P$ iff $\left(P^I\right)^\star = I$, i.e. if $I$ is the answer set of the reduct $P^I$.

### 2.2 Possibilistic Logic

At the semantic level, possibilistic logic is defined in terms of a *possibility distribution* $\pi$ on the universe of interpretations, i.e. an $\Omega \to [0,1]$ mapping which encodes for each interpretation, or possible world, $I$ to what extent it is plausible that $I$ is the actual world. Intuitively, $\pi(I)$ represents the compatibility of the interpretation $I$ with available information. By convention, $\pi(I) = 0$ means that $I$ is impossible and $\pi(I) = 1$ means that nothing prevents $I$ from being true in the real world. Note that possibility degrees are mainly interpreted qualitatively: when $\pi(I) > \pi(I')$, $I$ is considered more plausible than $I'$.

A possibility distribution $\pi$ induces two uncertainty measures. The *possibility measure* $\Pi$ is defined by

$$\Pi(p) = \max\{\pi(I) \mid I \models p\}$$

and evaluates the extent to which a proposition $p$ is consistent with the beliefs expressed by $\pi$. The dual

*necessity measure* $N$ is defined by

$$N(p) = 1 - \Pi(\neg p)$$

and evaluates the extent to which a proposition $p$ is entailed by the available beliefs (Dubois et al., 1994). An important property of necessity measures is min-decomposability: $N(p \wedge q) = \min(N(p), N(q))$ for all propositions $p$ and $q$.

At the syntactic level, a *possibilistic knowledge base* consists of pairs $(p, c)$ where $p$ is a proposition and $c \in [0, 1]$ expresses the certainty that $p$ is the case. A formula $(p, c)$ is interpreted as the constraint $N(p) \geq c$. A possibilistic knowledge base $\Sigma$ thus corresponds to a set of constraints on possibility distributions. Typically, there will be many possibility distributions that satisfy these constraints. In practice, we are usually only interested in those that make minimal commitments, called the least specific possibility distributions. Formally, a possibility distribution $\pi$ is a least specific possibility distribution compatible with $\Sigma$ if there is no possibility distribution $\pi'$ with $\pi' \neq \pi$ compatible with $\Sigma$, such that $\forall I \in \Omega \cdot \pi'(I) \geq \pi(I)$. Thus the least specific possibility distribution is the least informative possibility distribution.

In the remainder of this paper, we assume that all certainty weights are taken from a finite subset $C$ of $[0, 1]$.

### 2.3 Possibilistic ASP (PASP)

PASP combines ASP and possibility theory by associating a necessity value with atoms and rules. A *valuation* is a function $V : \mathcal{A} \to C$. The intuition is that for an atom $a \in \mathcal{A}$, $V(a) = c$ means that we can derive with certainty $c$ that is $a$ is true. For notational convenience, we also use the set notation $V = \{a^c, \ldots\}$. In accordance with this set notation, we write $V = \emptyset$ to denote the valuation in which each atom is mapped to 0. For $c \in C$ and a valuation $V$, we use $V^c$ to denote the classical interpretation $V^c = \{a \mid a \in \mathcal{A}, V(a) \geq c\}$. We use $V^{\underline{c}}$ to denote the classical interpretation $V^{\underline{c}} = \{a \mid a \in A, V(a) > c\}$, i.e. those atoms for which we can derive with certainty strictly greater than $c$ that they are true. A possibilistic normal (resp. definite) program is a set of pairs $p = (r, n(r))$ with $r$ a normal (resp. definite) rule and $n(r) \in C$ a certainty associated with $r$. We write a pair $p = (r, n(r))$ with $r = a_0 \leftarrow a_1, \ldots, a_m, \text{not } a_{m+1}, \ldots, \text{not } a_n$ as:

**n(r):** $a_0 \leftarrow a_1, \ldots, a_m, \text{not } a_{m+1}, \ldots, \text{not } a_n.$

The $c$-cut $P_c$ of a possibilistic normal program $P$ with $c \in C$ is defined as

$$P_c = \{(r, n(r)) \mid (r, n(r)) \in P; n(r) \geq c\}.$$

Similar as for classical definite programs, the unique answer set of a possibilistic definite program can be found using a syntactic transformation based on a fixpoint operator.

**Definition 1.** *Let $P$ be a possibilistic definite program and $V$ a valuation. The immediate consequence operator $T_P$ is defined for $c \in C$ as:*

$$T_P(V)(a_0) = \max\{c \mid V^c \models a_1, \ldots, a_m \\ \wedge (a_0 \leftarrow a_1, \ldots, a_m) \in P_c\}$$

The immediate consequence operator from Definition 1 is equivalent to the one proposed in (Nicolas et al., 2006), although our formulation is slightly different.

We use $P^\star$ to denote the fixpoint obtained by repeatedly applying $T_P$ starting from the minimal valuation $V = \emptyset$, i.e. the least fixpoint of $T_P$ w.r.t. set inclusion. A valuation $V$ is called an *answer set of a possibilistic definite program* if $V = P^\star$.

Answer sets of possibilistic normal programs are defined using a reduct. In (Nicolas et al., 2006) the reduct $P^A$ of a possibilistic normal program $P$ and a set of atoms $A$ is defined as

$$\begin{aligned} P^A = \{ & ((a_0 \leftarrow a_1, \ldots, a_m), n(r)) \mid (r, n(r)) \in P \\ & \wedge r = a_0 \leftarrow a_1, \ldots, a_m, \text{not } a_{m+1}, \ldots, \text{not } a_n \\ & \wedge \{a_{m+1}, \ldots, a_n\} \cap A = \emptyset\}. \end{aligned} \qquad (4)$$

A valuation $V$ is then called an *answer set of the possibilistic normal program* $P$ iff $\left(P^{(V^{\underline{0}})}\right)^\star = V$, i.e. if $V$ is the answer set of the reduct $P^{(V^{\underline{0}})}$.

## 3 Possibilistic semantics of ASP

In this section we introduce a new characterization of classical answer sets using possibility theory. Although many equivalent definitions have already been proposed (Lifschitz, 2008), studying alternative definitions often leads to new insights. In particular, as we will see in the next section, our possibilistic approach to ASP can be naturally generalized to PASP.

The core of our idea is to translate a normal program to a set of constraints on possibility distributions. The least specific possibility distributions that are compatible with these constraints will then correspond to the answer sets of the program. For example, a fact of the form '$a_0 \leftarrow$' expresses the constraint that only interpretations in which '$a_0$' is true should be possible. More precisely, the rule '$a_0 \leftarrow$' imposes the constraint $N(a_0) \geq N(\top)$ where $N(\top) = 1$ by definition, i.e. it is necessarily the case that $a_0$ is true. Similarly, a rule such as '$a_0 \leftarrow a_1$' translates to the constraint $N(a_0) \geq N(a_1)$: $a_0$ is at least as certain as $a_1$. Finally, a rule '$a_0 \leftarrow a_1, ..., a_m$' translates to $N(a_0) \geq N(a_1 \wedge ... \wedge a_m)$ which is equivalent to $N(a_0) \geq \min(N(a_1), ..., N(a_m))$ due to the min-decomposability property of necessity measures.

**Definition 2.** *Let $P$ be a definite program and $\pi : \Omega \to [0,1]$ a possibility distribution. For every $r \in P$, the constraint $\gamma(r)$ imposed by $r = (a_0 \leftarrow a_1, ..., a_m)$ is given by*

$$N(a_0) \geq \min(N(a_1), ..., N(a_m)).$$

Let $C_P = \{\gamma(r) \mid r \in P\}$ be the set of constraints imposed by program $P$. If $\pi$ satisfies the constraints in $C_P$, $\pi$ is said to be a possibilistic model of $C_P$, written $\pi \models C_P$. A possibilistic model of $C_P$ will also be called a possibilistic model of $P$. We write $S_P$ for the set of all least specific possibilistic models of $P$, i.e. for $\pi \in S_P$ there is no possibilistic model $\pi' \neq \pi$ of $P$ such that $\forall I \in \Omega \cdot \pi'(I) \geq \pi(I)$.

**Proposition 1.** *Let $P$ be a definite program. If $\pi \in S_P$ then $M = \{a \mid N(a) = 1, a \in \mathcal{B}_P\}$ is an answer set of $P$.*

**Proposition 2.** *Let $P$ be a definite program. If $M$ is an answer set of $P$ then the possibility distribution $\pi$ defined by $\pi(I) = 1$ iff $M \subseteq I$ and $\pi(I) = 0$ otherwise is contained in $S_P$.*

Note that because definite programs always have exactly one answer set, it follows that $S_P$ is a singleton.

Now we take a closer look at programs with negation-as-failure. Intuitively *not b* is true if we have no proof for $b$. However, the proof for $b$ may depend on other naf-atoms whose truth value, in turn, depends on $b$. To cope with this, the characterization of answer sets in terms of the Gelfond-Lifschitz reduct involves a guess: first, we guess for each naf-atom *not b* whether a proof for $b$ will be found and subsequently we verify whether our guess was correct. In our possibilistic approach, a similar strategy is possible. Specifically, for each naf-atom *not b* we guess with what certainty we will not find a proof for $b$, i.e. we guess the value of $1 - N(b) = \Pi(\neg b)$, for $N$ and $\Pi$ the necessity and possibility measures induced by a least specific possibilistic model of $P$. Intuitively, $g(b)$ reflects the certainty that "it is consistent to assume that $b$ is not true'.

**Definition 3.** *Let $P$ be a normal program and let $g$ be a mapping from $\mathcal{B}_P$ to $[0,1]$. For every $r \in P$, the constraint $\gamma_g(r)$ induced by $r = (a_0 \leftarrow a_1, ..., a_m, \text{not } a_{m+1}, ..., \text{not } a_n)$ and $g$ is given by*

$$N(a_0) \geq \min\left(N(a_1), ..., N(a_m), g(a_{m+1}), ..., g(a_n)\right). \tag{5}$$

$C_{(P,g)} = \{\gamma_g(r) \mid r \in P\}$ is the set of constraints imposed by program $P$ and $g$, and $S_{(P,g)}$ is the set of all least specific possibilistic models of $C_{(P,g)}$.

Before we prove the correspondence between the answer sets of $P$ and the possibility distributions in $S_{(P,g)}$, we illustrate the intuition in the following example.

**Example 1.** *Consider the program $P$ with the rules*

$$a \leftarrow \qquad b \leftarrow b \qquad c \leftarrow a, \text{not } b.$$

*The set of constraints $C_{(P,g)}$ is given by*

$$N(a) \geq N(\top) = 1$$
$$N(b) \geq N(b)$$
$$N(c) \geq \min\left(N(a), g(b)\right).$$

*We can rewrite the first constraint as $1 - \Pi(\neg a) \geq 1$ and thus $\Pi(\neg a) = 0$. The second constraint is always satisfied and can be dropped. The last constraint can be rewritten as $\Pi(\neg c) \leq 1 - \min(1 - \Pi(\neg a), g(b))$, which imposes an upper bound on the value that $\Pi(\neg c)$ can assume. This inequality can be further simplified since we already concluded that $\Pi(\neg a) = 0$. Hence we have $\Pi(\neg c) \leq 1 - \min(1 - 0, g(b)) = 1 - g(b)$. In conclusion, the program imposes the constraints*

$$\Pi(\neg a) = 0 \qquad \Pi(\neg c) \leq 1 - g(b).$$

*The set $S_{(P,g)}$ contains exactly one element, which is defined by*

$$\pi(\{a, b, c\}) = 1 \qquad \pi(\{b, c\}) = 0$$
$$\pi(\{a, b\}) = 1 - g(b) \qquad \pi(\{b\}) = 0$$
$$\pi(\{a, c\}) = 1 \qquad \pi(\{c\}) = 0$$
$$\pi(\{a\}) = 1 - g(b) \qquad \pi(\{\}) = 0.$$

*Note that this possibility distribution is independent of the choice for $g(a)$ and $g(c)$ since there are no occurrences of not $a$ and not $c$ in $P$. In particular, we can now establish for which choices of $g(b)$ it holds that $g(b) = \Pi(\neg b)$:*

$$g(b) = \Pi(\neg b) = \max\{\pi(I) \mid I \models \neg b\} = 1$$

*and thus we have that $\pi(\{a, b\}) = \pi(\{a\}) = 0$.*

*To find out whether $a$, $b$ and $c$ are necessarily true (i.e. belong to the corresponding answer set), we need to verify whether $N(a) = 1$, for $N$ the necessity measure induced by this particular least specific possibility distribution $\pi$. We have $N(a) = 1 - \Pi(\neg a) = 1$, $N(c) = 1 - \Pi(\neg c) = 1$ and $N(b) = 1 - \Pi(\neg b) = 0$. The unique answer set of $P$ is therefore $\{a, c\}$.*

**Proposition 3.** *Let $P$ be a normal program and let $g$ be a mapping from $\mathcal{B}_P$ to $[0,1]$. Let $\pi \in S_{(P,g)}$ be such that*

$$\forall a \in \mathcal{B}_P \cdot g(a) = \Pi(\neg a) \tag{6}$$

*and that*

$$\forall a \in \mathcal{B}_P \cdot N(a) \in \{0, 1\} \tag{7}$$

*then $M = \{a \mid N(a) = 1, a \in \mathcal{B}_P\}$ is an answer set of the normal program $P$.*

**Proposition 4.** *Let $P$ be a normal program. If $M$ is an answer set of $P$, there is a mapping $g$ from $\mathcal{B}_P$ to $[0, 1]$ and a possibility distribution $\pi \in S_{(P,g)}$ such that $\forall a \in \mathcal{B}_P \cdot g(a) = \Pi(\neg a)$, $\forall a \in M \cdot N(a) = 1$ and $\forall a \in (\mathcal{B}_P \setminus M) \cdot N(a) = 0$.*

## 4 Semantics of possibilistic ASP

In the previous section we have introduced a possibilistic semantics for ASP. These semantics can easily be extended to the case of PASP. Consider a rule of the form

**n(r):** $a_0 \leftarrow a_1, ..., a_m, \text{not } a_{m+1}, ..., \text{not } a_n$

we have already discussed that $a_0 \leftarrow a_1, ..., a_m, \text{not } a_{m+1}, ..., \text{not } a_n$ imposes a set of restrictions on the possibility distribution. Intuitively, the certainty of the head $a_0$ depends both on the certainty that the body is true and the certainty that the rule is actually valid. The former certainty can be calculated as for normal rules, and is given by the right-hand side of (5). The latter certainty is explicitly asserted to be $n(r)$. Because the necessity of a conjunctive statement is equal to the minimum of the necessities of its conjuncts, we arrive at the following semantics.

**Definition 4.** *Let $P$ be a possibilistic normal program and $g$ a mapping from $\mathcal{B}_P$ to $[0, 1]$. For every $p \in P$, the constraint $\gamma_g(p)$ induced by $p = (r, n(r))$ with $r = (a_0 \leftarrow a_1, ..., a_m, \text{not } a_{m+1}, ..., \text{not } a_n)$ and $g$ is given by*

$$N(a_0) \geq \min(N(a_1), ..., N(a_m),$$
$$g(a_{m+1}), ..., g(a_n), n(r)).$$

$C_{(P,g)} = \{\gamma_g(p) \mid p \in P\}$ *is the set of constraints imposed by program $P$, and $S_{(P,g)}$ is the set of all least specific possibilistic models of $C_{(P,g)}$.*

We can now easily generalize the characterization from Section 3 to define the semantics of possibilistic normal programs.

**Definition 5.** *Let $P$ be a possibilistic normal program and let $g$ be a mapping from $\mathcal{B}_P$ to $[0, 1]$. Let $\pi \in S_{(P,g)}$ be such that*

$$\forall a \in \mathcal{B}_P \cdot g(a) = \Pi(\neg a)$$

*then $V = \{a^{N(a)} \mid a \in \mathcal{B}_P\}$ is called a possibilistic answer set of $P$.*

This natural generalization of our characterization of classical answer sets precisely captures the intuition that 'not $a$' should be interpreted as "it cannot be established that $a$ is certain", as illustrated in the next example.

**Example 2.** *Consider program $P_1$ from Section 1. For compactness, we use $cB$ to denote concertBooked, $lD$ to denote longDrive and can to denote canceled. The set of constraints $C_{(P_1,g)}$ is given by*

$$N(cB) \geq \min(N(\top), 1)$$
$$N(lD) \geq \min(N(cB), g(can), 1)$$
$$N(can) \geq \min(N(\top), 0.2).$$

*We know that $N(\top) = 1$. Hence the first constraint can be rewritten as $N(cB) = 1$ or equivalently $\Pi(\neg cB) = 0$. Since $N(cB) = 1$, the second constraint can be simplified to $\Pi(\neg lD) \leq 1 - g(can)$. The last constraint can be simplified to $\Pi(\neg can) \leq 0.8$. The program thus imposes the constraints*

$$\Pi(\neg cB) = 0$$
$$\Pi(\neg lD) \leq 1 - g(can)$$
$$\Pi(\neg can) \leq 0.8.$$

*The set $S_{(P_1,g)}$ contains just one element defined by*

| | |
|---|---|
| $\pi(\{cB, lD, can\}) = 1$ | $\pi(\{lD, can\}) = 0$ |
| $\pi(\{cB, lD\}) = 0.8$ | $\pi(\{lD\}) = 0$ |
| $\pi(\{cB, can\}) = 1 - g(can)$ | $\pi(\{can\}) = 0$ |
| $\pi(\{cB\}) = \min(0.8, 1 - g(can))$ | $\pi(\{\}) = 0.$ |

*We can now establish for which choices of $g(can)$ it holds that $g(can) = \Pi(\neg can)$ :*

$$g(can) = \Pi(\neg can) = \max\{\pi(I) \mid I \models \neg can\} = 0.8$$

*and thus we have $\pi(\{cB, can\}) = \pi(\{cB\}) = 0.2$. We have $N(cB) = 1 - \Pi(\neg cB) = 1$, $N(lD) = 1 - \Pi(\neg lD) = 1 - 0.2 = 0.8$ and $N(can) = 1 - \Pi(\neg can) = 1 - 0.8 = 0.2$. The unique possibilistic answer set of $P_1$ is therefore $\{cB^1, lD^{0.8}, can^{0.2}\}$.*

It is important to note that under the semantics proposed in this paper there is no longer a 1-on-1 mapping between the classical answer sets of a normal program and the possibilistic answer sets of that same normal program with certainty 1 attached to each of the rules. Indeed, comparing Definition 5 with Proposition 3, we can see that no equivalent to (7) is imposed for possibilistic answer sets.

**Example 3.** *Consider the normal program with the single rule $a \leftarrow \text{not } a$. This program has no classical answer sets. Now consider the possibilistic normal program $P$ with the rule*

**1:** $a \leftarrow \text{not } a.$

The set of constraints $C_{(P,g)}$ is given by

$$N(a) \geq \min(g(a), 1).$$

This constraint can be rewritten as $\Pi(\neg a) \leq 1 - g(a)$. The set $S_{(P,g)}$ contains just a single element defined by $\pi(\{a\}) = 1$ and $\pi(\{\}) = 1 - g(a)$. We can now establish for which choices of $g(a)$ it holds that $g(a) = \Pi(\neg a)$:

$$g(a) = \Pi(\neg a) = 1 - g(a)$$

and thus $\pi(\{\}) = 0.5$. The unique possibilistic answer set of $P$ is therefore $\{a^{0.5}\}$. This should come as no surprise. In classical ASP we are limited to conclusions with absolute certainty, a limitation that we do not have in PASP. In the same way, one may verify that the program

    **1:** $a \leftarrow not\ b$          **1:** $b \leftarrow not\ a$

has an infinite number of possibilistic answer sets, i.e. $\{a^c, b^{1-c}\}$ for every $c \in [0,1]$.

## 5 Possibilistic reduct

In the previous section, we have defined answer sets for PASP at the semantic level, in terms of possibility distributions satisfying certain constraints. This definition is a natural generalization of the characterization of answer sets in ASP from Section 3, but quite different from how answer sets are traditionally defined, namely through the Gelfond-Lifschitz reduct. In this section, we close this gap by introducing a generalization of the Gelfond-Lifschitz reduct, leading to a purely syntactic procedure for finding possibilistic answer sets. In contrast to (4), our reduct is defined w.r.t. a valuation and not simply a set of atoms. Hence, the certainty weights are taken into account when determining the reduct of a possibilistic normal program.

**Definition 6.** *Let $V$ be a valuation. The reduct $P^V$ of a possibilistic normal program $P$ is defined as:*

$$P^V = \{((a_0 \leftarrow a_1, ..., a_m), \min(c_1, c_2)) \mid \min(c_1, c_2) > 0$$
$$\land c_2 = \max\{c \mid \{a_{m+1}, ..., a_n\} \cap V^{1-c} = \emptyset, c \in [0,1]\}$$
$$\land ((a_0 \leftarrow a_1, ..., a_m, not\ a_{m+1}, ..., not\ a_n), c_1) \in P\}.$$

Note that $P^V$ is a possibilistic definite program.

**Example 4.** *Consider the program $P_1$. Let $V$ be a valuation such that $V = \{cB^1, lD^{0.8}, can^{0.2}\}$. The first rule of $P_1$ is found unchanged in $(P_1)^V$ since we only need to consider the body $\top$ which clearly has no naf-atoms. Hence $\min(c_1, c_2) = 1$ with $c_1 = c_2 = 1$ and thus $(cB \leftarrow, 1) \in (P_1)^V$. A similar line of though can be followed for the last rule, though this rule can only be found as $(can \leftarrow, 0.2) \in P$. Hence $\min(0.2, 1) = 0.2$ and we have that $(can \leftarrow, 0.2) \in (P_1)^V$. For the second rule, which has a non-empty body, we find that $\{can\} \cap V^{1-c} = \emptyset$ for $c \leq 0.8$. We obtain that $\min(1, 0.8) = 0.8$ and thus that $(lD \leftarrow cB, 0.8) \in (P_1)^V$.*

Interestingly, the reduct we obtain here is equivalent to the reduct that has been proposed in (Madridás and Aciego, 2008) for residuated logic programs, in the special case where conjunction is modeled using the minimum. Both approaches are different in spirit, however, in the same way that possibilistic logic (which deals with uncertainty or priority) is different from Gödel logic (which deals with graded truth). This formal correspondence with residuated logic programs is due to the fact that necessity measures are min-decomposable. As possibilistic logic is not truth-functional in general, it is easy to see that possibilistic ASP diverges from residuated logic programming as soon as disjunction and strict negation are considered.

Before we can show how to find possibilistic answer sets using this reduct, we need to determine the syntactic least fixpoint operator that corresponds with our proposed semantics. For definite programs, it is easy to see that our semantics is equivalent to the one from (Nicolas et al., 2006), hence the least fixpoint operator from Definition 1 can be used to this end.

**Proposition 5.** *Let $P$ be a possibilistic normal program. If $V$ is an answer set of $P$ then $V = (P^V)^\star$.*

**Proposition 6.** *Let $P$ be a possibilistic normal program and $V$ a valuation. If $V = (P^V)^\star$ then $V$ is an answer set of $P$.*

**Example 5.** *Once again consider the possibilistic program $P_1$ from Section 1. Let $V$ be a valuation such that $V = \{cB^1, lD^{0.8}, can^{0.2}\}$. We can now easily verify that $V$ is an answer set of $P_1$. From Example 4 we know that*

$$(P_1)^V = \{(cB \leftarrow, 1), (lD \leftarrow cB, 0.8), (can \leftarrow, 0.2)\}$$

*for which it is easy to verify that $\left((P_1)^V\right)^\star = V$. We can also easily verify that $V' = \{cB^1, can^{0.2}\}$ is not an answer set. We obtain the same reduct as before and clearly $\left((P_1)^{V'}\right)^\star = \left((P_1)^V\right)^\star \neq V'$.*

## 6 Simulation with classical ASP

The definition of the reduct from the previous section suggests a technique to simulate PASP using classical ASP. This is important, as it allows to straightforwardly implement PASP using existing ASP solvers.

Specifically, a possibilistic normal program $P$ is simulated by the normal program $Q$ containing the rules

$$\{a_0c' \leftarrow a_1c', ..., a_mc', not\ a_{m+1}c'', ..., not\ a_nc''\ |$$
$$((a_0 \leftarrow a_1, ..., a_m, not\ a_{m+1}, ..., not\ a_n), c) \in P,$$
$$0 < c' \leq c, c' \in C,$$
$$c'' = \min\{d\ |\ d > 1 - c', d \in C\}\}$$

where we introduce a fresh atom $ac$ for each $a$ in the original program and $c \in C$.

**Example 6.** *Consider $P_1$ from Section 1. For compactness, let $C = \{0, 0.2, 0.4, 0.6, 0.8, 1\}$. The classical normal program $Q_1$ that simulates $P_1$ contains the rules*

$$cB(1) \leftarrow \quad cB(0.8) \leftarrow \quad cB(0.6) \leftarrow$$
$$cB(0.4) \leftarrow \quad cB(0.2) \leftarrow \quad can(0.2) \leftarrow$$

$$lD(1) \leftarrow cB(1), not\ can(0.2)$$
$$lD(0.8) \leftarrow cB(0.8), not\ can(0.4)$$
$$lD(0.6) \leftarrow cB(0.6), not\ can(0.6)$$
$$lD(0.4) \leftarrow cB(0.4), not\ can(0.8)$$
$$lD(0.2) \leftarrow cB(0.2), not\ can(1.0).$$

$Q_1$ *has the answer set*

$$M = \{cB(1), cB(0.8), cB(0.6), cB(0.4), cB(0.2)\}$$
$$\cup \{lD(0.8), lD(0.6), lD(0.4), lD(0.2)\} \cup \{can(0.2)\}$$

*which indeed corresponds to the possibilistic answer set $\{cB^1, lD^{0.8}, can^{0.2}\}$ of $P_1$.*

The next two propositions confirm that this simulation is indeed correct and that we can use a classical normal program to obtain the possibilistic answer sets of our possibilistic normal program.

**Proposition 7.** *Let $P$ be a possibilistic normal program and $Q$ the simulation of $P$ defined above. Let $V$ be the valuation defined by $V(a) = \max\{c\ |\ ac \in M\}$. If $M$ is a classical answer set of $Q$, then $V$ is a possibilistic answer set of $P$.*

**Proposition 8.** *Let $P$ be a possibilistic normal program and $Q$ the simulation of $P$ defined above, and $M = \{ac\ |\ c \leq V(a) = c', c \in C\}$. If $V$ is a possibilistic answer set of $P$ such that $V(a) \in C$ for all $a \in \mathcal{B}_P$, then $M$ is a classical answer set of $Q$.*

Only those possibilistic answer sets for which $V(a) \in C$ are found using our simulation. In particular, thus only a finite number of answer sets can be found.

## 7 Related Work

A large body of research has been devoted to combining ASP with uncertainty. This uncertainty can either be interpreted in a qualitative or in a quantitative way. In the latter case, probability theory is most often used. For example, in (Baral et al., 2009) uncertainty is encoded by means of probability atoms. Intuitively, a probability atom describes the probability that the atom will take on a certain value in some random selection given other known evidence.

Instead of probability theory, some approaches use evidence theory as the underlying model of uncertainty. For instance, (Wan, 2009) uses belief functions to allow intervals of certainty degrees to be attached to rules, instead of a single value. A similar approach can already be found in Fril (Baldwin et al., 1995) where fuzzy set theory is employed to derive certainty degrees from linguistic variables.

The most popular approach to deal with uncertainties in a qualitative fashion is possibility theory. Adopting possibility theory in logic programming was an idea pioneered in (Dubois et al., 1991), although default negation was not considered in this early work. One of the first papers to explore this idea in the context of ASP was (Nicolas et al., 2006) in which the authors present a framework that combines possibility theory with ASP. This approach was later extended to disjunctive programs (Nieves et al., 2007). Alternative semantics for PASP have been proposed based on pstable models (Confalonieri et al., 2009). However, pstable models (Osorio et al., 2006) are closer to classical models than they are to stable models, i.e. answer sets. The semantics based on pstable models capture a different intuition, focusing more on finding reasonable results in the face of inconsistency. For instance, the program containing the rule **c:** $a \leftarrow not\ a$ has $(a, c)$ as its unique possibilistic pstable model, which is not compatible with a reading of '*not a*' as "it cannot be established that $a$ is certain".

The constraint-based view on ASP we take in this paper is somewhat reminiscent of the idea of using SAT-based solvers to find answer sets (Lin and Zhao, 2002). Our approach also has some similarities to the one presented in (Benferhat et al., 1997) where default rules induce constraints on the possibility distribution. Specifically, a default rule "if $a$ then $b$" is interpreted as $\Pi(a \wedge b) > \Pi(a \wedge \neg b)$, which captures the intuition that when $a$ is known to hold, $b$ is more plausible than $\neg b$, unless information to the contrary becomes available. Similarly, they define entailment by looking at the least specific possibility distributions (although the notion of least specific possibility distribution is defined, in this context, w.r.t. the plausibility ordering on interpretations induced by the possibility degrees). Our approach also has some resemblance with the characterization of answer sets in auto-epistemic logic. Indeed, an ASP rule such as $a \leftarrow not\ b$ can be interpreted by the formula $\neg \mathsf{L}b \to a$ in auto-epistemic logic, where $\mathsf{L}b$ reflects the intuition "it is not believed (certain) that $b$" underlying our semantics.

From an applications perspective, the purpose of combining ASP with uncertainty theories is often to deal with inconsistencies. Handling inconsistencies is by itself a well-studied field within the ASP community. Of particular interest are approaches that deal with preferences. Indeed, in possibilistic logic necessity degrees are commonly interpreted as preferences between rules. One example of preferences in classical ASP is (Nieuwenborgh and Vermeir, 2002), where ordered logic programs are used. Ordered logic programs assume a partial order among rules, allowing less important rules to be violated in order to satisfy rules with higher importance. In some sense, the use of such preferences among rules is related to using certainty weights, although the resulting semantics is closer in spirit to the approach from (Nicolas et al., 2006) than to the semantics we have developed in this paper. Quite a number of other works also deal with preference handling in non-monotonic reasoning; we refer to (Delgrande et al., 2004) for a good overview.

## 8 Conclusions

We have introduced a new characterization of classical answer sets using possibility theory. Under these semantics, rules in a classical normal program impose a set of constraints on possibility distributions. The classical answer sets then correspond with the least specific possibilistic distributions that satisfy these constraints and adhere to some general restrictions. Subsequently, we have extended this idea to cover possibilistic ASP. The central observation was that the certainty of the conclusion of a rule should be given by the certainty that both the body of the rule is valid and that the rule itself is valid. We have also introduced a syntactic reduct that precisely corresponds to the proposed semantics for PASP. Finally, to demonstrate the applicability of our approach, we have shown how possibilistic normal programs can be simulated by classical normal programs.

**Acknowledgements**

The authors would like to thank the anonymous reviewers for their useful comments, in particular for pointing out the link between our work and residuated logic programs.